\documentclass[sigplan,nonacm]{acmart}

\usepackage{amsmath}
\usepackage{amsfonts}
\usepackage{booktabs}
\usepackage{graphicx}
\usepackage{multirow}
\usepackage{multicol}
\usepackage{tablefootnote}
\usepackage{float}
\usepackage{url}
\usepackage{makecell}
\usepackage{hyperref}
\usepackage{algorithm}
\usepackage{algpseudocode}
\AtBeginDocument{%
  }

\setcopyright{acmlicensed}
\copyrightyear{2018}
\acmYear{2018}
\acmDOI{XXXXXXX.XXXXXXX}
\acmConference[BIOKDD 2026]{25th International Workshop on Data Mining in Bioinformatics}{August 9-13, 2026}{Woodstock, NY}
\acmISBN{978-1-4503-XXXX-X/2018/06}

\begin{document}


\title{PRIME: Protein Representation via Physics-Informed Multiscale Equivariant Hierarchies}

\author{Viet Thanh Duy Nguyen}
\affiliation{
  \institution{University of Alabama at Birmingham}
  \city{Birmingham}
  \state{Alabama}
  \country{USA}
}
\email{dvnguye2@uab.edu}

\author{John K. Johnstone}
\affiliation{
  \institution{University of Alabama at Birmingham}
  \city{Birmingham}
  \state{Alabama}
  \country{USA}
}
\email{jkj@uab.edu}

\author{Truong-Son Hy}
\authornote{Corresponding Author}
\affiliation{
  \institution{University of Alabama at Birmingham}
  \city{Birmingham}
  \state{Alabama}
  \country{USA}
}
\email{thy@uab.edu}

\renewcommand{\shortauthors}{Nguyen et al.}




\begin{abstract}
Proteins are inherently multiscale physical systems whose functional properties emerge from coordinated structural organization across multiple spatial resolutions, ranging from atomic interactions to global fold topology. However, existing protein representation learning methods typically operate at a single structural level or treat different sources of structural information as parallel modalities, without explicitly modeling their hierarchical relationships. We introduce PRIME (Protein Representation via Physics-Informed Multiscale Equivariant Hierarchies), a unified framework that models proteins as a nested family of five physically grounded structural graphs spanning surface, atomic, residue, secondary-structure, and protein levels. Adjacent levels are connected through deterministic, physics-informed assignment operators, enabling bidirectional information exchange via bottom-up aggregation and top-down contextual refinement. Experiments on standard protein representation learning benchmarks demonstrate strong and competitive performance across diverse tasks, with particularly notable gains on the Fold Classification benchmark, where PRIME outperforms the strongest geometric GNN baseline by margins of 13.80 and 18.30 points on the harder Superfamily and Fold splits, and achieves a state-of-the-art accuracy of 84.10\% on Reaction Class prediction, surpassing all baseline methods, including ESM. Ablation studies confirm that each structural level contributes complementary and non-redundant information, and adaptive cross-attention analysis reveals that PRIME autonomously identifies the most task-relevant structural resolutions at prediction time.
Our source code is publicly available at \url{https://github.com/HySonLab/PRIME}.
\end{abstract}

\maketitle

\section{Introduction}

Proteins are inherently hierarchical physical systems whose structural and functional properties emerge from coordinated organization across multiple scales, from atomic interactions and local surface geometry to secondary structure elements and global fold topology. This multiscale organization is not merely a descriptive convenience: it reflects fundamental physical and biochemical principles, consistent with the framework of multiscale modeling in complex chemical systems recognized by the Nobel Prize in Chemistry 2013 \cite{karplus2014development}, where leveraging known physical organization across scales is precisely what distinguishes principled multiscale approaches from purely data-driven ones. Learning meaningful representations of such systems therefore requires modeling interactions that arise both within and across these structural scales. However, most existing protein representation learning methods operate at a single structural resolution, focusing on one level of description at a time, such as amino acid sequence \cite{lin2023evolutionary, hayes2025simulating}, three-dimensional atomic or residue-level structure \cite{wang2022point, guo2022self, zhang2023protein}, or molecular surface structure \cite{gainza2020deciphering, sverrisson2021fast}. Recent work has sought to address this limitation by incorporating multiple sources of structural information within multimodal learning frameworks \cite{lee2024pretraining, nguyen2024multimodal, xu2024surface}. However, these approaches typically model different structural representations as parallel modalities and integrate them through latent alignment or late-stage fusion. In doing so, distinctions between atomic geometry, residue organization, and surface morphology are treated as independent semantic views, rather than as nested levels within a shared physical hierarchy. As a result, existing multimodal methods do not explicitly account for the hierarchical organization of the protein structure, limiting their ability to capture interactions that emerge across structural resolutions.

A more faithful approach is to explicitly model the hierarchical organization of the protein structure rather than collapsing distinct spatial resolutions through post hoc alignment. Recent work in graph representation learning has explored hierarchical graph constructions and multiresolution message passing as a means of capturing long-range dependencies and structured interactions \cite{hy2023multiresolution, ngo2023multiresolution, trang2024scalable}. These approaches typically learn graph coarsenings directly from data and have shown promise in generic domains. However, in protein systems, the relevant hierarchy from atoms to residues to secondary-structure elements and global folds is explicitly defined by physical and biochemical principles, providing a natural and principled inductive bias for hierarchical graph representation learning over multiscale geometric systems.

Motivated by these considerations, we introduce PRIME (Protein Representation via Physics-Informed Multiscale Equivariant Hierarchies). This hierarchical graph representation learning framework models proteins through a nested family of physically grounded structural graphs spanning multiple spatial resolutions. Here, the term \textit{physics-informed} does not imply that physical constraints are incorporated into the training objective, as is conventional in physics-informed neural networks, but rather reflects that the hierarchy itself is grounded in the known multiscale physical organization of proteins, with each level and its connecting operators derived from established biochemical relationships rather than inferred from data. PRIME enables information exchange across resolutions through deterministic assignment operators derived from protein structure, while leveraging pretrained equivariant encoders at the finest-grained levels to produce geometrically rich invariant features from 3D atomic and surface structure. By learning over this hierarchical graph structure, PRIME captures interactions both within and across structural scales in a coherent multiscale representation. This formulation provides a principled foundation for protein representation learning and complements existing unimodal and multimodal approaches by explicitly accounting for the hierarchical organization of protein structure.

Our main contributions are as follows.
\begin{itemize}
    \item We propose PRIME (Protein Representation via Physics-Informed Multiscale Equivariant Hierarchies), a unified hierarchical graph representation learning framework that models proteins as a nested family of five physically grounded structural graphs, connected via deterministic, physics-informed assignment operators that enable bidirectional information exchange across resolutions through bottom-up aggregation and top-down contextual refinement.
    \item We evaluate PRIME on standard protein representation learning benchmarks, demonstrating strong and competitive performance across diverse tasks. Ablation studies and adaptive cross-attention analysis further confirm that each structural level contributes complementary information and that the model autonomously identifies the most task-relevant structural resolutions at prediction time.
\end{itemize}

\section{Related Work}

\paragraph{Protein Representation Learning}
Protein representation learning has been investigated across several structural resolutions, including sequence-based models that capture evolutionary patterns \cite{lin2023evolutionary, hayes2025simulating}, structure-based methods operating on atomic or residue-level graphs \cite{wang2022point, guo2022self, zhang2023protein}, and surface-based approaches that encode molecular geometry and physicochemical properties relevant to molecular recognition \cite{gainza2020deciphering, sverrisson2021fast}. Given the complementary nature of these representations, recent work has explored multimodal frameworks that jointly incorporate sequence, structure, and surface information \cite{lee2024pretraining, nguyen2024multimodal, xu2024surface}. However, these approaches typically treat different structural resolutions as parallel modalities and integrate them through latent alignment or late-stage fusion. As a result, existing multimodal methods do not explicitly model the hierarchical physical organization of the protein structure, operating over flat representations rather than nested structural levels.

\paragraph{Hierarchical Graph Representation Learning.} Hierarchical graph representation learning methods aim to capture long-range dependencies and multiscale structure by constructing representations across multiple resolutions through graph pooling, clustering, or coarsening operations that aggregate nodes into higher-level abstractions \cite{hy2023multiresolution, ngo2023multiresolution, trang2024scalable}. Such multiresolution schemes have demonstrated strong performance in capturing structured relationships across large-scale and complex graph domains. However, most existing frameworks rely on data-driven partitioning strategies to infer graph hierarchies during training, which may not align with the underlying physical organization of structured geometric systems. In proteins, hierarchical relationships between molecular surfaces, atoms, residues, secondary-structure elements, and global folds are explicitly defined by biochemical and geometric constraints, providing a natural and principled basis for physics-informed hierarchical graph construction without requiring uncertain data-driven clustering.

\section{Preliminaries}
\label{sec:preliminaries}

A structured system can be modeled as a nested hierarchy of graphs capturing interactions across multiple spatial resolutions. Let $\mathcal{G} = \{ G^{(0)}, \dots, G^{(L)} \}$ denote a family of graphs ordered from fine ($\ell=0$) to coarse ($\ell=L$) resolution, where each $G^{(\ell)} = (V^{(\ell)}, E^{(\ell)})$ consists of a node set $V^{(\ell)}$ and an edge set $E^{(\ell)}$ encoding structural relationships at resolution $\ell$. Node features at level $\ell$ are represented as a matrix $Z^{(\ell)} \in \mathbb{R}^{|V^{(\ell)}| \times d_\ell}$.

The hierarchy is constructed such that each coarse-level unit corresponds to a collection of elements from the preceding finer level, formalized through assignment maps $\pi^{(\ell)}: V^{(\ell-1)} \rightarrow V^{(\ell)}$ for $\ell = 1, \dots, L$. Each assignment map is equivalently represented by a binary partition matrix $\Pi^{(\ell)} \in \{0,1\}^{|V^{(\ell-1)}| \times |V^{(\ell)}|}$, where $\Pi^{(\ell)}_{ik} = 1$ if fine-scale node $i \in V^{(\ell-1)}$ is assigned to coarse node $k \in V^{(\ell)}$ and $0$ otherwise. Each matrix $\Pi^{(\ell)}$ induces a disjoint partition of fine-scale nodes into coarse structural units.

The connectivity of $G^{(0)}$ is represented by a binary adjacency matrix $A^{(0)} \in \{0,1\}^{|V^{(0)}| \times |V^{(0)}|}$. Adjacency matrices at coarser levels are derived deterministically via:
\[
A^{(\ell)} = (\Pi^{(\ell)})^\top A^{(\ell-1)} \Pi^{(\ell)}, \quad \ell = 1, \dots, L,
\]
yielding weighted adjacency matrices where $A^{(\ell)}_{ij}$ reflects the number of fine-scale connections between the structural units of coarse nodes $i$ and $j$. Together, $\{ Z^{(\ell)} \}$, $\{ A^{(\ell)} \}$, and $\{ \Pi^{(\ell)} \}$ collectively define a multiresolution graph hierarchy over nested levels of structural organization.

\section{Methodology}

\begin{figure*}[t]
    \centering
    \includegraphics[width=\linewidth]{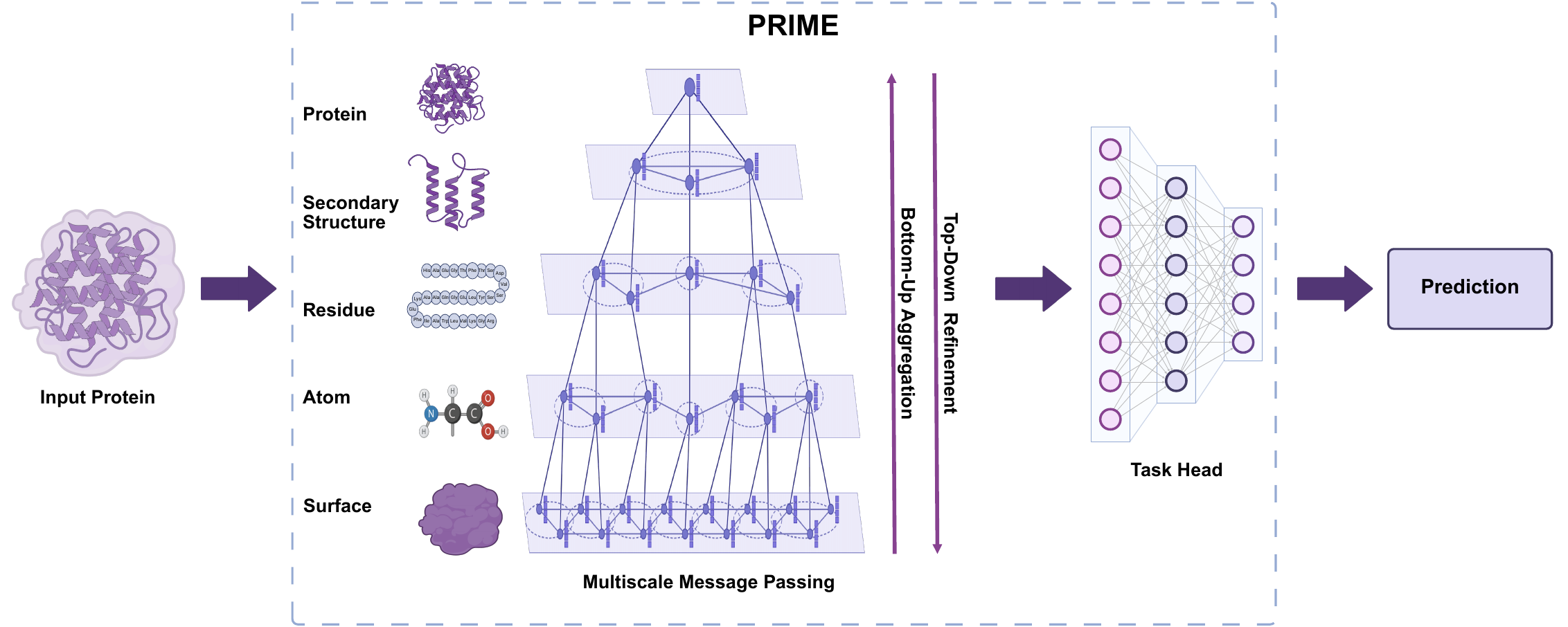}
    \caption{Overview of the PRIME framework. An input protein is represented as a nested hierarchy of five physically grounded graphs spanning the surface, atomic, residue, secondary-structure, and protein levels. Cross-scale interactions are propagated bidirectionally through bottom-up aggregation and top-down refinement. The resulting protein-level representation is passed to a task head for downstream prediction.}
    \label{fig:prime}
\end{figure*}

PRIME, as illustrated in Figure \ref{fig:prime}, represents each input protein as a nested hierarchy of five physically grounded graphs, spanning from fine-grained surface and atomic structure to coarser residue, secondary-structure, and protein-level organization. The framework proceeds in two stages. First, the protein is decomposed into a multiresolution graph hierarchy, where each level captures distinct structural and physicochemical information through modality-specific node features, and adjacent levels are connected through deterministic, physics-informed assignment matrices derived from known biochemical relationships (Section \ref{sec:graph_construction}). Second, representations are learned through a multiscale message-passing scheme that propagates information both within each level via intra-level graph convolutions and across levels via bidirectional cross-scale fusion, combining bottom-up aggregation of fine-grained detail with top-down contextual refinement from global structure (Section \ref{sec:message_passing}). The final representation at a designated readout level is passed to a task-specific head for downstream prediction. Full implementation details, including model architecture, optimization, and training infrastructure, are provided in Appendix \ref{app:implementation}, and a formal analysis of the time and space complexity of PRIME is provided in Appendix \ref{app:complexity}.

\subsection{Protein Hierarchical Graph Construction}
\label{sec:graph_construction}

Building on the multiresolution graph framework introduced in Section \ref{sec:preliminaries}, PRIME represents each protein as a nested hierarchy of graphs that explicitly reflects the multiscale physical organization of the protein structure. Formally, the hierarchy is defined as:
\[
\mathcal{G} = \{ G^{(0)}, G^{(1)}, G^{(2)}, G^{(3)}, G^{(4)} \},
\]
spanning five physically grounded levels of structural organization ordered from fine to coarse resolution.

\paragraph{Node sets.} Each graph $G^{(\ell)}$ is defined over a distinct set of structural units: $V^{(0)}$ consists of triangular faces of the molecular surface mesh; $V^{(1)}$ consists of heavy atoms representing fine-grained three-dimensional atomic structure; $V^{(2)}$ consists of amino acid residues encoding backbone geometry; $V^{(3)}$ consists of secondary-structure elements (SSEs) including helices, strands, and loops; and $V^{(4)}$ consists of a single node representing the entire protein.

\paragraph{Edge sets.} Edges at the finest level $G^{(0)}$ are constructed via $k$-nearest neighbors ($k=8$) over face centroids, connecting each surface face to its spatially closest neighbors. Edges at coarser levels are derived deterministically from $A^{(0)}$ through successive application of the partition matrices:
\[
A^{(\ell)} = (\Pi^{(\ell)})^\top A^{(\ell-1)} \Pi^{(\ell)}, \quad \ell = 1, \dots, 4,
\]
yielding weighted adjacency matrices where each entry $A^{(\ell)}_{ij}$ reflects the number of fine-scale connections between the structural units of coarse nodes $i$ and $j$. The resulting matrices are subsequently normalized using symmetric degree normalization:
\[
\hat{A}^{(\ell)} = D^{-1/2} A^{(\ell)} D^{-1/2},
\]
where $D$ is the diagonal degree matrix with $D_{ii} = \sum_j A^{(\ell)}_{ij}$, and rows with zero degree are left unchanged. This normalization prevents high-degree nodes from dominating message aggregation, bounds the eigenvalues of $\hat{A}^{(\ell)}$ for training stability, and ensures consistent message scales across proteins of varying size, while preserving the symmetry of the adjacency matrix for undirected message passing. While common approaches construct independent graphs at each level using $k$-nearest neighbors or radius-based connectivity in 3D coordinate space, our coarsening-based approach derives edge weights that naturally encode spatial proximity; two coarse nodes with many fine-scale connections between their constituent elements are necessarily spatially close, providing a geometrically grounded connectivity structure without requiring explicit coordinate-based edge construction at each level.

\paragraph{Node Features.} At each level $\ell$, nodes are associated with modality-specific feature vectors $Z^{(\ell)} \in \mathbb{R}^{|V^{(\ell)}| \times d_\ell}$. At the surface level, features comprise two handcrafted geometric descriptors (face area and distance to the protein centroid) concatenated with invariant scalar embeddings from a pretrained Equivariant Mesh Neural Network (EMNN) \cite{trang20243}. At the atomic level, features are invariant scalar embeddings from a pretrained Equivariant Graph Neural Network (EGNN) \cite{satorras2021n}, which takes element type and backbone membership as input. Both encoders are pretrained by us on protein structural data; full details are provided in Appendix \ref{app:pretraining}. At the residue level, features comprise a one-hot encoding of amino acid identity, one backbone bond angle, one local curvature estimate, and one torsional dihedral angle. At the secondary-structure level, each node is represented by a one-hot encoding of its structural type (helix, strand, or loop) and a normalized length scalar. At the protein level, a global representation is obtained from a pretrained ESM-2 650M model \cite{lin2023evolutionary}, encoding evolutionary and functional information for the entire sequence.

\paragraph{Assignment Maps.} Adjacent levels are connected through 
deterministic assignment maps $\pi^{(\ell)}: V^{(\ell-1)} \rightarrow 
V^{(\ell)}$, derived from known physical and geometric relationships 
rather than learned from data. These operators fall into two categories. The first are \textit{physically-grounded} mappings: atoms are mapped to their parent residue via the covalent bonding structure; each residue is assigned to its secondary-structure segment via the DSSP algorithm \cite{kabsch1983dictionary}; and secondary-structure elements are aggregated into a single protein-level node. These mappings are uniquely defined by established physical and chemical principles of protein organization. The second is a \textit{geometrically-motivated} mapping: each surface face is assigned to its nearest atom by Euclidean distance from its centroid, a well-established convention in molecular surface modeling that provides a principled geometric interface between the two finest-grained levels of the hierarchy. Full implementation details of the partition matrix construction are provided in Appendix 
\ref{app:partition_matrix_construction}.

\subsection{Multiscale Message Passing}
\label{sec:message_passing}

Given the hierarchical protein graph $\mathcal{G}$, PRIME learns representations by propagating information both within and across structural levels through a stack of $N$ hierarchical layers. Each layer performs three sequential operations, intra-level message passing, bottom-up aggregation, and top-down contextual refinement, as summarized in Algorithm \ref{alg:prime}. Node features at each level $\ell$ are first projected to a shared hidden dimension $d$ via a level-specific linear transformation, followed by layer normalization and ReLU activation. Each hierarchical layer proceeds through three sequential operations. \textit{Intra-level message passing} first refines representations at each level independently through a graph convolution block that applies pre-layer normalization, separate linear transformations for self and neighbor aggregation over $A^{(\ell)}$, dropout, a residual connection, and a two-layer feedforward network with GELU activation. \textit{Bottom-up aggregation} then propagates fine-grained structural detail upward by aggregating fine-scale representations into coarse nodes via the transpose partition matrices $(\Pi^{(\ell)})^\top$, followed by a level-specific linear projection $W^{(\ell)}_{\uparrow}$ and gated fusion into the coarse-level features. \textit{Top-down contextual refinement} completes the layer by broadcasting coarse-level context back to finer levels via the forward partition matrices $\Pi^{(\ell)}$, projected by a level-specific linear layer $W^{(\ell)}_{\downarrow}$ and fused symmetrically through the same gating mechanism. A residual connection is applied across hierarchical layers: denoting the input to layer $n$ at level $\ell$ as $H^{(\ell)}_{n-1}$ and the output as $H^{(\ell)}_n$, the final representation is $H^{(\ell)}_n \leftarrow H^{(\ell)}_n + H^{(\ell)}_{n-1}$, facilitating gradient flow during training.

A key property of this bidirectional message passing scheme is that after $N$ layers, the representation at every level $\ell$ encodes not only its own modality-specific information but also contextual signals from all other levels, fine-grained geometric detail propagated upward through bottom-up aggregation, and global structural context propagated downward through top-down refinement. This makes each level's representation a holistic summary of the protein at its corresponding structural resolution. At each cross-scale interaction, the gated fusion mechanism selectively incorporates context $c$ from an adjacent level into the current representation $z$:
\begin{align*}
g = \sigma(W_g \, [z \| c]), \quad u = W_u \, c, \quad z \leftarrow \text{LayerNorm}(z + g \odot u),
\end{align*}
where $[\cdot \| \cdot]$ denotes concatenation, $\sigma$ denotes sigmoid activation, $\odot$ is element-wise multiplication, $W_g \in \mathbb{R}^{d \times 2d}$ and $W_u \in \mathbb{R}^{d \times d}$ are learnable weight matrices, and dropout is applied to the update $u$ for regularization. After $N$ layers, the representation $H^{(\ell^*)}_N$ at a designated readout level $\ell^*$ is passed to a task-specific prediction head for downstream prediction. A formal analysis of the SE(3) invariance properties of the final representations is provided in Appendix \ref{app:invariance}.

\begin{algorithm}[t]
\caption{PRIME: Multiscale Message Passing}
\label{alg:prime}
\begin{algorithmic}[1]
\Require Hierarchical graph $\mathcal{G} = \{G^{(0)}, \dots, G^{(4)}\}$, partition matrices $\{\Pi^{(\ell)}\}_{\ell=1}^{4}$, number of layers $N$, readout level $\ell^*$
\Ensure Task representation $H^{(\ell^*)}_N$
\For{$\ell = 0$ \textbf{to} $4$}
    \State $H^{(\ell)}_0 \leftarrow \text{ReLU}(\text{LayerNorm}(W^{(\ell)}_{\text{in}} Z^{(\ell)}))$ \Comment{Input projection}
\EndFor
\For{$n = 1$ \textbf{to} $N$}
    \State \textbf{// Intra-Level Message Passing}
    \For{$\ell = 0$ \textbf{to} $4$}
        \State $H^{(\ell)}_n \leftarrow H^{(\ell)}_{n-1} + W_{\text{self}} H^{(\ell)}_{n-1} + W_{\text{neigh}} A^{(\ell)} H^{(\ell)}_{n-1}$ \Comment{Graph convolution}
        \State $H^{(\ell)}_n \leftarrow H^{(\ell)}_n + \text{FFN}(H^{(\ell)}_n)$ \Comment{Feedforward block}
    \EndFor
    \State \textbf{// Bottom-Up Aggregation}
    \For{$\ell = 1$ \textbf{to} $4$}
        \State $\tilde{H}^{(\ell)}_n \leftarrow W^{(\ell)}_{\uparrow} \left( (\Pi^{(\ell)})^\top H^{(\ell-1)}_n \right)$
        \State $H^{(\ell)}_n \leftarrow \text{GatedFusion}(H^{(\ell)}_n,\, \tilde{H}^{(\ell)}_n)$
    \EndFor
    \State \textbf{// Top-Down Contextual Refinement}
    \For{$\ell = 4$ \textbf{downto} $1$}
        \State $\tilde{H}^{(\ell-1)}_n \leftarrow W^{(\ell)}_{\downarrow} \left( \Pi^{(\ell)} H^{(\ell)}_n \right)$
        \State $H^{(\ell-1)}_n \leftarrow \text{GatedFusion}(H^{(\ell-1)}_n,\, \tilde{H}^{(\ell-1)}_n)$
    \EndFor
    \State $H^{(\ell)}_n \leftarrow H^{(\ell)}_n + H^{(\ell)}_{n-1}$ \text{ for all } $\ell$ \Comment{Inter-layer residual}
\EndFor
\State \Return $H^{(\ell^*)}_N$ \Comment{Readout from designated level $\ell^*$}
\end{algorithmic}
\end{algorithm}

\section{Experiments}

\subsection{Benchmarks for Protein Representation Learning}

We evaluate PRIME on four well-established protein representation learning benchmarks from ProteinWorkshop \cite{jamasb2024evaluating}, covering both graph-level and node-level prediction tasks. We directly use the preprocessed datasets and standard splits provided by ProteinWorkshop. For graph-level tasks, we consider Fold Prediction \cite{hou2018deepsf}, a multiclass classification task evaluated on three test splits of increasing difficulty (Family, Superfamily, and Fold); Reaction Class Prediction \cite{hermosilla2021intrinsicextrinsic}, a multiclass classification task predicting the enzyme commission class of a protein; and Gene Ontology Prediction \cite{gligorijevic2021structure}, a multilabel classification task over three functional ontologies, Biological Process (BP), Molecular Function (MF), and Cellular Component (CC). For node-level tasks, we evaluate on Protein-Protein Interaction (PPI) Site Prediction \cite{gainza2020deciphering}, a binary residue-level classification task predicting interaction interface residues. We follow the standard evaluation metrics from ProteinWorkshop: accuracy for Fold and Reaction Class, $F_{\max}$ for Gene Ontology, and ROC-AUC for PPI Site Prediction. For each baseline, we report the best result across all featurisation schemes and auxiliary tasks from ProteinWorkshop \cite{jamasb2024evaluating}, with the exception of ESM, for which we report results without structural feature augmentation to provide a pure sequence-based reference for comparison. Detailed descriptions of each task and dataset statistics are provided in Appendix \ref{app:datasets}.

\subsection{PRIME Performance Evaluation}

We evaluate PRIME across a diverse set of protein downstream tasks to assess the generality and effectiveness of its multiscale hierarchical representations. We fix the readout level to $\ell^* = \text{residue}$ across all tasks to ensure a fair and consistent comparison with baseline methods that predominantly operate at residue resolution. This choice is further justified by PRIME's architectural design: the residue level occupies a natural middle position in the hierarchy, capturing both local geometric detail from finer levels and global structural context from coarser levels through bidirectional message passing, making it well-suited for the broad range of tasks considered. For the node-level PPI task, residue-level readout is the most natural choice; for graph-level tasks, residue features are mean-pooled to obtain a protein-level representation for prediction. As shown in Table \ref{tab:main_results}, PRIME achieves strong and competitive performance across all benchmarks, demonstrating the effectiveness of explicitly modeling the hierarchical physical organization of protein structure for learning generalizable protein representations.

On the Fold Classification benchmark, PRIME achieves the highest accuracy across all three test splits, outperforming EGNN, the strongest geometric GNN baseline, by substantial margins of 13.80 and 18.30 points on the Superfamily and Fold splits, respectively. This is particularly notable on the harder splits where sequence similarity to training proteins is low, suggesting that the multiscale hierarchy enables PRIME to capture structural features that generalize beyond sequence-level similarity. The complementary contributions of SSE-level topology and fine-grained atomic and residue-level geometry may be key to discriminating structurally similar folds at these harder generalization levels.

On the Reaction Class prediction benchmark, PRIME surpasses all baselines with an accuracy of 84.10, including ESM, which benefits from large-scale pretraining on hundreds of millions of protein sequences. This suggests that explicit structural hierarchy provides a complementary and in some cases superior signal to sequence-level representations for enzyme function prediction. Reaction class is governed by the active site, where catalytic properties emerge from the precise interplay of surface geometry, local atomic chemical environments, and backbone geometry, factors that span multiple structural resolutions and are most naturally captured jointly.

On the Gene Ontology prediction benchmark, PRIME achieves the best performance on the MF and CC ontologies, while underperforming on the BP ontology. The strong performance on MF and CC, which are more directly tied to local structural motifs and global fold properties, respectively, suggests that multiscale structural representations are particularly beneficial for ontologies with strong structure-function coupling. The lower performance on BP likely reflects the broader and more heterogeneous nature of biological process annotations, which encode high-level cellular and metabolic processes that may depend more on evolutionary context and interaction partners than on intrinsic structural properties, where fine-grained structural signals may contribute limited discriminative information or even introduce noise.

On the PPI Site Prediction benchmark, PRIME achieves competitive but not top-ranked ROC-AUC on this node-level task. While interaction interfaces are shaped by the joint interplay of surface geometry, local atomic properties, and higher-order structural context, the relatively lower performance suggests that the precise local geometric characterization required for residue-level interface prediction may benefit from more specialized architectural choices. Nevertheless, the competitive performance demonstrates that PRIME's multiscale representations are broadly applicable to both graph-level and node-level prediction tasks.

\begin{table*}[ht]
\centering
\caption{Performance comparison on protein representation learning benchmarks. Metrics: accuracy for Fold and Reaction Class, $F_{\max}$ for Gene Ontology (GO), and ROC-AUC for PPI. Best results are in \textbf{bold}.}
\label{tab:main_results}
\resizebox{\linewidth}{!}{
\begin{tabular}{lcccccccc}
\toprule
\multirow{2}{*}{Method} & \multicolumn{3}{c}{Fold Classification} & \multicolumn{3}{c}{Gene Ontology} & \multirow{2}{*}{Reaction} & \multirow{2}{*}{PPI} \\
\cmidrule(lr){2-4} \cmidrule(lr){5-7}
& Family & Superfamily & Fold & BP & MF & CC & & \\
\midrule
ESM \cite{lin2023evolutionary} & 97.80 & 60.10 & 26.80 & \textbf{0.462} & 0.546 & 0.394 & 83.10 & 0.955 \\
\midrule
SchNet \cite{schutt2018schnet}     & 90.31 & 37.66 & 29.48 & 0.343 & 0.417 & 0.429 & 73.83 & 0.956 \\
GearNet \cite{zhang2023protein}    & 95.48 & 48.39 & 34.63 & 0.404 & 0.483 & 0.453 & 80.03 & 0.962 \\
EGNN \cite{satorras2021n}          & 97.29 & 56.29 & 41.48 & 0.373 & 0.494 & 0.455 & 82.70 & 0.965 \\
GCPNet \cite{morehead2024geometry} & 96.55 & 51.85 & 38.86 & 0.371 & 0.470 & 0.442 & 77.71 & \textbf{0.968} \\
TFN \cite{thomas2018tensor}        & 96.09 & 52.98 & 36.65 & 0.375 & 0.489 & 0.452 & 80.84 & 0.967 \\
MACE \cite{batatia2022mace}        & 95.94 & 48.27 & 37.05 & 0.350 & 0.457 & 0.411 & 76.10 & 0.965 \\
\midrule
PRIME (Ours) & \textbf{98.11} & \textbf{70.09} & \textbf{59.78} & 0.412 & \textbf{0.578} & \textbf{0.456} & \textbf{84.10} & 0.957 \\
\bottomrule
\end{tabular}
}
\end{table*}

\subsection{Ablation Study: Effect of Hierarchical Levels}

To understand the contribution of each structural level to PRIME's performance, we conduct an ablation study by systematically removing individual levels from the active hierarchy while keeping all other components fixed. Importantly, removing a level also removes its associated pretrained encoder: excluding the surface level removes the EMNN encoder, excluding the atomic level removes the EGNN encoder, and excluding the protein level removes the ESM-2 embedding. The ablation, therefore, simultaneously evaluates both the contribution of each structural resolution and its associated pretrained feature extractor. When a level is excluded, its intra-level GNN and all associated cross-scale transitions are bypassed, resulting in a disconnected computational graph at that level. This is intentional: rather than re-wiring conducted on the Fold Classification benchmark, ensure the strict contribution of each level by removing both its representations and its role as a communication bridge within the hierarchy. This design reflects the true cost of losing a structural level in PRIME, capturing not only the loss of its modality-specific features but also its topological role in enabling cross-scale information flow. The readout level is fixed to $\ell^* = \text{residue}$ throughout so that observed performance changes can be attributed solely to the removal of each level, and ablations are performed on the Fold Classification benchmark due to computational constraints.

As shown in Table \ref{tab:ablation_levels}, removing any single level consistently degrades performance across all three splits, confirming that each structural resolution contributes complementary and non-redundant information. The most pronounced degradation occurs upon removal of the SSE level, with accuracy dropping from 59.78 to 16.61 on the hardest Fold split, underscoring the central role of secondary structure organization in fold discrimination. This is consistent with biological intuition, as fold identity is fundamentally defined by the global spatial arrangement of helices and strands, making SSE-level representations the most directly informative for distinguishing protein folds. The protein level shows the second largest drop, reflecting the importance of global sequence context from the ESM-2 embedding; however, this drop is considerably smaller than that from SSE removal, indicating that PRIME does not rely predominantly on sequence-level features. The significant degradation upon removing purely structural levels, namely surface, atom, and SSE, further demonstrates that PRIME's gains arise from joint multiscale structural modeling rather than from any single pretrained encoder alone. These results confirm that the full five-level hierarchy is necessary for strong performance and that no single level or its associated encoder is redundant. The relative contribution of each level is task-dependent, and this task-dependence motivates PRIME's unified multiscale design, in which all structural resolutions are jointly maintained and mutually reinforced through bidirectional message passing, ensuring that task-relevant information at any resolution is preserved without requiring manual level selection.

\begin{table*}[ht]
\centering
\caption{Ablation on hierarchical levels on Fold Classification. Each column removes one structural level from the full PRIME hierarchy.}
\label{tab:ablation_levels}
\resizebox{\linewidth}{!}{
\begin{tabular}{lccccc}
\toprule
Split & PRIME (full) & w/o Surface & w/o Atom & w/o SSE & w/o Protein \\
\midrule
Family      & \textbf{98.11} & 86.54 & 91.03 & 56.18 & 74.24 \\
Superfamily & \textbf{70.09} & 55.07 & 62.55 & 31.33 & 52.92 \\
Fold        & \textbf{59.78} & 40.73 & 51.70 & 16.61 & 36.14 \\
\bottomrule
\end{tabular}
}
\end{table*}

\subsection{Multiscale Readout via Cross-Attention}

A key advantage of PRIME is that it simultaneously produces a rich representation at every structural level after message passing, rather than committing to a single resolution. While the ablation study examines level contributions through controlled removal, here we take a complementary internal perspective: we ask which levels the model itself chooses to rely on when given unconstrained access to all representations at the final prediction stage. To investigate this, we introduce an adaptive cross-attention readout in which a single learnable query vector attends over the mean-pooled representations from all five levels, producing a unified task embedding that dynamically weights each structural resolution. We evaluate this on the Fold Classification benchmark and analyze the learned attention weights to gain insight into the model's structural preferences at prediction time. To further examine whether level preferences vary across tasks, additional results on the Reaction Class benchmark are provided in Appendix \ref{app:additional_results}.

\begin{figure*}[ht]
    \centering
    \includegraphics[width=\linewidth]{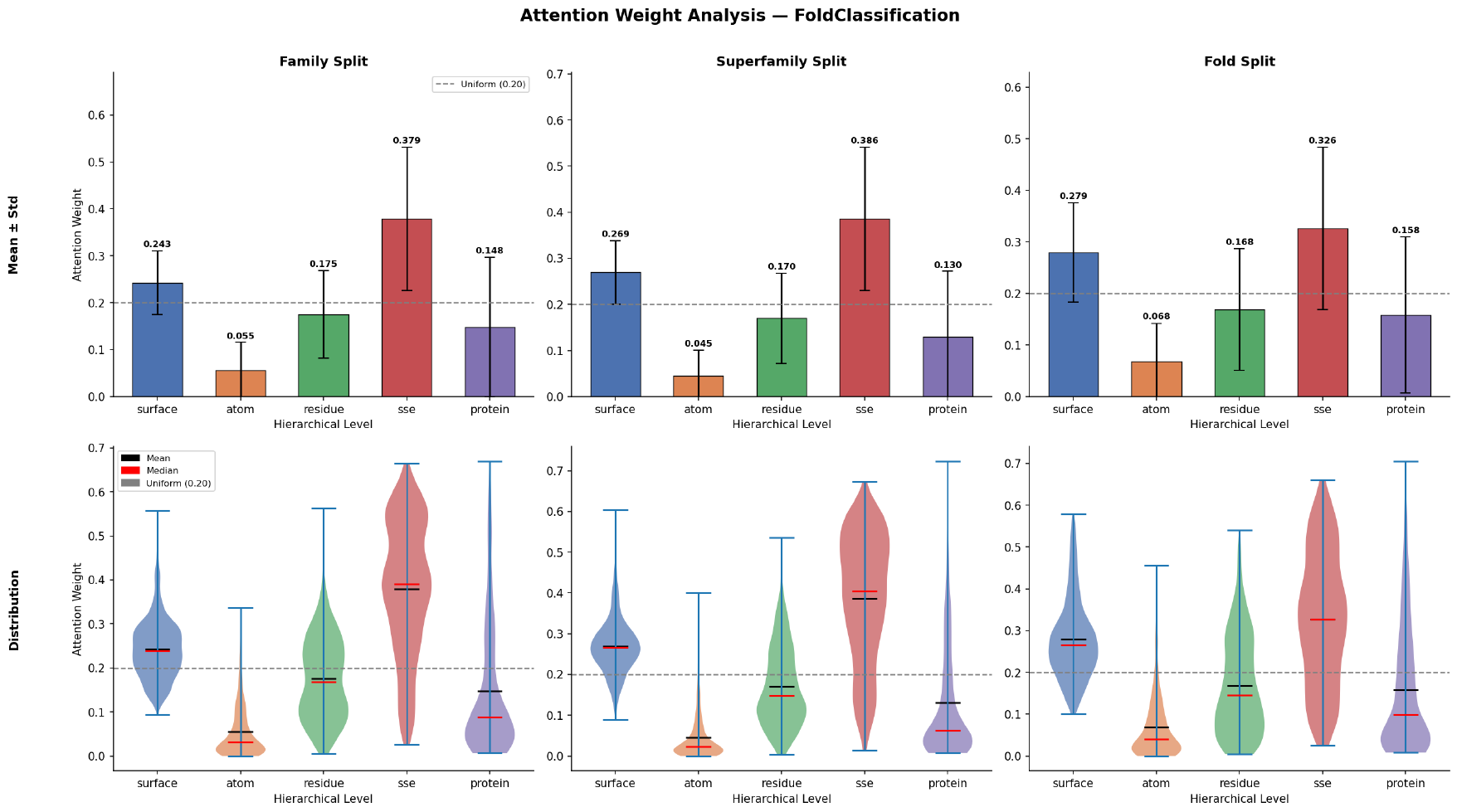}
    \caption{Learned attention weights over structural levels for the adaptive cross-attention readout on Fold Classification.}
    \label{fig:attn_combined}
\end{figure*}

The adaptive cross-attention readout achieves accuracy of \textbf{98.81, 70.73, and 60.22} on the Family, Superfamily, and Fold splits, respectively, slightly surpassing the fixed residue-level readout across all three splits, suggesting that aggregating representations across all hierarchical levels provides richer task-relevant information than relying on a single fixed resolution. As shown in Figure \ref{fig:attn_combined}, the SSE level receives the highest attention weight, consistent with fold identity being fundamentally defined by the global spatial arrangement of helices and strands, while the atom level is suppressed well below the uniform baseline as individual atomic positions provide a limited discriminative signal for fold-level classification. Surface and residue representations receive moderate attention, and the protein-level embedding is underweighted, suggesting that intermediate structural resolutions are more informative than global sequence context for fold discrimination. The attention patterns are consistent across all three splits, while the violin plots reveal meaningful per-protein variance, suggesting the model dynamically adjusts its reliance on different levels. Notably, the ranking of level importance is consistent with the ablation findings, providing converging evidence from two independent perspectives that PRIME autonomously identifies the most task-relevant structural information at prediction time without requiring manual specification of the readout level. We note that attention weights reflect the model's preference at the readout stage and do not preclude indirect contributions of low-weight levels through inter-level message passing.

\section{Conclusion}

We presented PRIME, a hierarchical graph representation learning framework that models proteins as a nested family of five physically grounded structural graphs spanning surface, atomic, residue, secondary-structure, and protein levels. By propagating information bidirectionally across levels through physics-informed assignment operators and gated cross-scale fusion, PRIME captures interactions both within and across structural resolutions within a unified multiscale representation. Experiments on standard protein representation learning benchmarks demonstrate strong and competitive performance across diverse tasks, confirming the effectiveness of explicitly modeling the hierarchical physical organization of protein structure for learning generalizable protein representations. Ablation studies confirm that each structural level contributes complementary and non-redundant information, with the full hierarchy outperforming any partial configuration. Furthermore, the adaptive cross-attention readout analysis reveals that PRIME autonomously identifies the most task-relevant structural resolutions at prediction time, with learned attention weights consistent with the ablation findings and aligned with biological intuition.

Several limitations of the current work suggest promising directions for future investigation. First, PRIME currently employs deterministic, physics-informed assignment operators that are fixed during training; investigating soft, learnable assignment operators initialized from physical priors but refined during training could capture task-specific structural groupings beyond canonical biochemical definitions. Second, the framework currently models individual protein chains in isolation; extending it to protein complexes by introducing intermolecular interaction levels would broaden its applicability to tasks such as binding affinity prediction and docking. Third, coarser-level adjacency matrices are currently derived solely from the surface mesh; exploring hybrid adjacency constructions that combine surface-derived edges with canonical structural edges such as chemical bonds or radius-based proximity graphs could further enrich the connectivity structure of the hierarchy, providing complementary relational information beyond what is captured by the surface-coarsening approach alone. Fourth, the equivariant encoders are pretrained on a relatively modest dataset due to storage constraints from high-resolution surface meshes; scaling pretraining to larger structural corpora, or exploring more storage-efficient surface representations such as point clouds, remains an important direction. Finally, PRIME is currently trained end-to-end in a supervised manner for each downstream task; exploring self-supervised pretraining objectives across the full hierarchical graph, such as cross-scale contrastive learning or multiscale denoising, could enable it to learn rich multiscale representations from large unlabeled protein databases, reducing reliance on labeled data.

\bibliographystyle{plain}
\bibliography{references}

\appendix

\section{Implementation Details}
\label{app:implementation}

\paragraph{Model Architecture.} PRIME is implemented in PyTorch. All levels are projected to a shared hidden dimension of $d = 128$ via level-specific linear transformations, and the hierarchical GNN consists of $N = 3$ layers with dropout rate $p = 0.3$ applied throughout both the graph convolution blocks and the gated fusion modules. Input feature dimensionalities are determined by each level's representation: 130 for surface (128-dimensional EMNN encoder embeddings concatenated with 2 handcrafted geometric descriptors comprising face area and distance to centroid), 128 for atom (invariant scalar embeddings from the pretrained EGNN encoder), 23 for residue (20-dimensional one-hot amino acid identity, one backbone bond angle, one local curvature estimate, and one torsional dihedral angle), 4 for secondary-structure (3-dimensional one-hot structural type and one normalized length scalar), and 1280 for protein (ESM-2-650M CLS token). The task-specific prediction head is a three-layer MLP with hidden dimension 128 and dropout $p = 0.3$, shared in architecture across all tasks.

\paragraph{Optimization.} All models are trained using the AdamW optimizer with a learning rate of $\eta = 10^{-3}$ and a weight decay of $\lambda = 10^{-4}$, with a batch size of 32. A linear warmup schedule is applied over the first 3 epochs, ramping the learning rate from $0.01\eta$ to $\eta$. Following warmup, a \texttt{ReduceLROnPlateau} scheduler reduces the learning rate by a factor of 0.6 when the validation metric does not improve for 5 consecutive epochs. Gradient norms are clipped to a maximum of 5.0 to stabilize training. Models are trained for up to 100 epochs, with early stopping triggered after 10 epochs of no improvement in validation performance, and the best checkpoint is selected based on the highest validation metric.

\paragraph{Loss Functions.} For multi-label classification tasks (Gene Ontology prediction), binary cross-entropy with logits loss (\texttt{BCEWithLogitsLoss}) is used. For single-label classification tasks (Fold Classification and Reaction Class prediction), standard cross-entropy loss (\texttt{CrossEntropyLoss}) is used. For node-level tasks (PPI Site Prediction), binary cross-entropy with logits loss is used with sigmoid activation applied during evaluation.

\paragraph{Training Infrastructure.} All experiments are conducted on a single NVIDIA A100-SXM4 80GB GPU. Peak GPU memory usage varies by task, reaching up to approximately 56GB for the most memory-intensive benchmark, PPI Site Prediction, due to its large number of residue-level nodes.

\section{Complexity Analysis}
\label{app:complexity}

\paragraph{Time Complexity.} We analyze the time complexity of PRIME per forward pass, where $d$ denotes the shared hidden dimension and $N$ denotes the number of hierarchical layers. The complexity consists of three components. For intra-level message passing at level $\ell$, each graph convolution involves both sparse message aggregation over $A^{(\ell)}$ with complexity $\mathcal{O}(|E^{(\ell)}| \cdot d)$ and dense node-wise linear transformations ($W_{\text{self}}$, $W_{\text{neigh}}$, and the two-layer feedforward network) with complexity $\mathcal{O}(|V^{(\ell)}| \cdot d^2)$. The total intra-level complexity per layer at level $\ell$ is therefore $\mathcal{O}((|E^{(\ell)}| + |V^{(\ell)}| \cdot d) \cdot d)$, and summing over all five levels and $N$ layers gives:
\[
\mathcal{O}\left(N \cdot \sum_{\ell=0}^{4} \left( |E^{(\ell)}| \cdot d + |V^{(\ell)}| \cdot d^2 \right)\right).
\]

For cross-scale message passing, bottom-up aggregation via $(\Pi^{(\ell)})^\top H^{(\ell-1)}$ has complexity $\mathcal{O}(|V^{(\ell-1)}| \cdot d)$ per transition, since $\Pi^{(\ell)}$ is a sparse binary matrix with exactly $|V^{(\ell-1)}|$ non-zero entries, followed by a linear projection of complexity $\mathcal{O}(|V^{(\ell)}| \cdot d^2)$. Similarly, top-down refinement has complexity $\mathcal{O}(|V^{(\ell-1)}| \cdot d + |V^{(\ell-1)}| \cdot d^2)$ per transition. The total cross-scale complexity per layer is therefore $\mathcal{O}(\sum_{\ell=1}^{4} (|V^{(\ell-1)}| + |V^{(\ell)}|) \cdot d^2)$.

We note that the assumption $|V^{(0)}| \geq |V^{(1)}| \geq \dots \geq |V^{(4)}|$ holds in general by the hierarchical construction, where each coarse level aggregates multiple fine-scale nodes. However, in practice, the surface and atomic levels are subject to independent size caps: surface meshes are decimated to at most 1,024 faces, while atomic graphs are subsampled to at most 2,048 atoms. For large proteins that exceed both thresholds, it is therefore possible that $|V^{(1)}| > |V^{(0)}|$, which formally violates the monotonicity assumption. In such cases, the theoretical bound does not strictly hold, and the dominant term is $\mathcal{O}(|V^{(1)}| \cdot d^2)$ rather than $\mathcal{O}(|V^{(0)}| \cdot d^2)$. Nevertheless, since both node counts are bounded by fixed constants in practice (1,024 and 2,048, respectively), the overall complexity remains bounded by:
\[
\mathcal{O}\left(N \cdot \left( |E^{(0)}| \cdot d + C \cdot d^2 \right)\right),
\]
where $C = \max(|V^{(0)}|, |V^{(1)}|) \leq 2048$ is a fixed constant determined by the preprocessing caps, making PRIME computationally tractable regardless of protein size.

\paragraph{Space Complexity.} The space complexity of PRIME consists of two components: model parameters and intermediate activations. The number of trainable parameters is $\mathcal{O}(N \cdot L \cdot d^2)$, where $N$ is the number of hierarchical layers and $L = 5$ is the number of levels, arising from the level-specific input projections, graph convolution weights, and gated fusion parameters at each level and layer. Since the parameters are not shared across layers, each of the $N$ layers maintains its own set of $\mathcal{O}(L \cdot d^2)$ parameters, yielding a total parameter count that scales linearly with both the number of layers and levels. The memory required to store intermediate node representations across all levels is $\mathcal{O}(\sum_{\ell=0}^{4} |V^{(\ell)}| \cdot d)$, dominated by the finest levels. The partition matrices $\{\Pi^{(\ell)}\}$ are stored as sparse matrices with $\mathcal{O}(\sum_{\ell=1}^{4} |V^{(\ell-1)}|)$ non-zero entries in total, contributing negligible memory overhead. Since all node counts are bounded by fixed constants after preprocessing, the overall space complexity per protein is $\mathcal{O}(C \cdot d + N \cdot L \cdot d^2)$, where $C$ is the maximum node count across all levels.

\section{Equivariant Encoder Pretraining Details}
\label{app:pretraining}

Both equivariant encoders, the Equivariant Mesh Neural Network (EMNN) for surface-level features and the Equivariant Graph Neural Network (EGNN) for atomic-level features, are pretrained independently on protein structural data using a self-supervised denoising autoencoder objective. In both cases, only the encoder weights are retained after pretraining and used as fixed feature extractors within PRIME.

\paragraph{Pretraining Dataset.} Both encoders are pretrained on a curated dataset of 11,258 non-redundant protein structures with paired atomic coordinates and triangular surface meshes, originally curated for the MaSIF framework \cite{gainza2020deciphering}. Each entry consists of a PDB structure and its corresponding molecular surface mesh, enabling joint pretraining of both encoders on the same set of proteins. We note that the dataset size is primarily constrained by storage requirements, as storing high-resolution triangular surface meshes is considerably more expensive than storing atomic coordinates alone; scaling to larger pretraining corpora such as AlphaFoldDB \cite{fleming2025alphafold} remains an important direction for future work. We note that potential overlap between the pretraining corpus and benchmark test sets does not constitute data leakage for two reasons. First, the encoders are used as fixed feature extractors and are not fine-tuned during downstream training, meaning no task-specific information can be absorbed after pretraining. Second, the pretraining objective is purely self-supervised, predicting corrupted coordinates from noisy inputs, with no access to downstream task labels or benchmark splits, making it structurally incapable of encoding task-relevant information regardless of structural overlap. This setup is directly analogous to the use of pretrained ESM-2 embeddings, which have been exposed to the vast majority of known protein sequences including benchmark test proteins, yet are widely accepted as a fair baseline in the protein representation learning community.

\paragraph{Pretraining Objective.} Both encoders are trained under a denoising autoencoder framework. Given input 3D coordinates $x$, Gaussian noise is added to obtain corrupted coordinates $\tilde{x} = x + \epsilon$, where $\epsilon \sim \mathcal{N}(0, \sigma^2 I)$ and $\sigma \sim \mathcal{U}(0.1, 0.5)$ is sampled uniformly per batch. The encoder maps the corrupted input to a latent representation $z$, and a lightweight three-layer MLP decoder with SiLU activations predicts the residual $x - \tilde{x}$ from $z$. The training objective is a mean squared error loss over the predicted residual:
\[
    \mathcal{L}_{\text{denoise}} = \left\| f_\theta(\tilde{x}) - (x - \tilde{x}) \right\|^2,
\]
augmented with a smoothness regularization term over the encoder's latent representations:
\[
    \mathcal{L}_{\text{smooth}} = \frac{1}{|E|} \sum_{(i,j) \in E} \left\| f_\theta(\tilde{x})_i - f_\theta(\tilde{x})_j \right\|^2.
\]

We note that this term does not model the noise targets directly, since the injected noise is sampled i.i.d. per node, encouraging smooth noise predictions would be mathematically inconsistent. Instead, $\mathcal{L}_{\text{smooth}}$ serves as a regularizer on the encoder's latent space, encouraging geometrically consistent representations across neighboring nodes in the graph structure, independent of the denoising objective. The total training loss is:
\[
    \mathcal{L} = \mathcal{L}_{\text{denoise}} + 0.01 \cdot \mathcal{L}_{\text{smooth}}.
\]

Models are optimized using the Adam optimizer with a cosine annealing learning rate schedule and gradient clipping to a maximum norm of 1.0. The dataset is split 90/10 into training and validation sets, and the best encoder checkpoint is selected based on validation loss. After pretraining, the decoder is discarded, and only the encoder $f_\theta$ is retained as a fixed feature extractor within PRIME.

\paragraph{EGNN Pretraining.} The EGNN encoder is pretrained on atomic-level protein graphs constructed from PDB structures. Hydrogen atoms are excluded, and heavy atoms are represented by a six-dimensional one-hot encoding of their element type. Atomic coordinates are normalized prior to training, and edges are constructed via $k$-nearest neighbors ($k=8$) in the 3D coordinate space with Euclidean distances as edge attributes. Proteins exceeding 2048 atoms are randomly subsampled to ensure tractable graph sizes, consistent with prior work on atomic protein graph learning \cite{nguyen2024multimodal}.

\paragraph{EMNN Pretraining.} The EMNN encoder is pretrained on molecular surface graphs derived from triangular mesh representations of protein surfaces. Meshes exceeding 1024 faces are first simplified via quadric decimation to ensure uniform graph sizes, following established practice in mesh-based learning \cite{feng2019meshnet}. Each triangular face is treated as a node, with its centroid serving as the node position, and edges are defined by mesh adjacency between neighboring faces.

\section{Partition Matrix Construction}
\label{app:partition_matrix_construction}

\paragraph{Molecular Surface Generation.} Protein molecular surfaces are generated from PDB structures using PyMOL \cite{PyMOL} with a solvent radius of 1.4 \AA, producing triangular mesh representations of the solvent-excluded molecular surface. This solvent radius corresponds to the approximate radius of a water molecule, making the generated surface physically meaningful as it represents the boundary accessible to solvent molecules and is therefore directly relevant to protein-ligand interactions, binding site geometry, and surface-mediated biological processes. Meshes are subsequently simplified to a maximum of 1,024 faces via quadric decimation implemented in Open3D \cite{zhou2018open3d}, following established practice in mesh-based learning \cite{feng2019meshnet, singh2021meshnet++} and ensuring memory-efficient graph sizes across proteins of varying length.

\paragraph{Surface-to-Atom Assignment.} Each triangular face is assigned to its nearest heavy atom based on the Euclidean distance from the face centroid, computed efficiently using a $k$-d tree over heavy atom coordinates. Hydrogen atoms are excluded from consideration. This assignment is physically grounded in the fact that the molecular surface geometry is primarily determined by the positions of heavy atoms, which define the van der Waals envelope of the protein. Assigning each surface face to its nearest heavy atom, therefore, establishes a natural geometric correspondence between the continuous surface representation and the discrete atomic structure underlying it.

\paragraph{Atom-to-Residue Assignment.} Each heavy atom is assigned to its parent amino acid residue according to the covalent bonding structure of the protein. This assignment is unambiguous and physically exact, as every heavy atom in a protein belongs to exactly one amino acid residue by definition of the peptide chain topology. The resulting partition reflects the fundamental chemical unit of protein structure, where residues serve as the basic building blocks from which higher-order structural organization emerges.

\paragraph{Residue-to-SSE Assignment.} Secondary-structure element assignments are determined using the DSSP algorithm \cite{kabsch1983dictionary} via the \texttt{pydssp} library, applied directly to the full set of residues from the protein structure prior to any atom subsampling. This ensures that backbone integrity is fully preserved during SSE assignment and that the resulting segmentation reflects the true secondary structure of the protein. Each maximal contiguous run of residues sharing the same DSSP label is treated as a single SSE node, yielding a deterministic segmentation of the protein chain into helices, strands, and loops. This grouping is physically motivated by the fact that secondary-structure elements represent cooperative hydrogen-bonding patterns along the backbone that give rise to structurally and functionally coherent units, making them a natural coarse-graining of the residue-level representation.

\paragraph{SSE-to-Protein Assignment.} All SSE nodes are aggregated into a single protein-level node, representing the global structural context of the entire protein chain. This final aggregation captures the overall fold topology of the protein, encoding the collective arrangement of all secondary-structure elements into a single holistic representation. The single protein-level node serves as a global context vector that, through top-down refinement, propagates information about the overall structural organization back to finer levels of the hierarchy.

\section{SE(3) Invariance Analysis}
\label{app:invariance}

We analyze the SE(3) symmetry properties of PRIME and show that the full framework achieves SE(3) invariance end-to-end, meaning that rotating or translating the input protein structure produces identical final representations.

\paragraph{Node Feature Invariance.} All input node features across all five hierarchical levels are SE(3)-invariant by construction. At the surface level, handcrafted features comprise face area and distance to the protein centroid, both scalar quantities invariant to rigid body transformations, concatenated with invariant scalar embeddings from the pretrained EMNN encoder. At the atomic level, node features are the invariant scalar embeddings produced by the pretrained EGNN encoder, which takes element type and backbone membership indicators as input but outputs only invariant scalar representations, discarding the equivariant coordinate outputs. At the residue level, features comprise a one-hot encoding of amino acid identity, one backbone bond angle, one local curvature estimate, and one torsional dihedral angle, all of which are scalar quantities invariant to rotation and translation. At the SSE level, features are a one-hot encoding of structural type and a scalar normalized length, both invariant. At the protein level, the ESM-2 embedding is derived purely from the amino acid sequence without any 3D coordinate information, and is therefore invariant.

\paragraph{Equivariant Encoders as Invariant Feature Extractors.} The EMNN and EGNN encoders are E(3)-equivariant by construction, meaning their internal message passing exploits 3D coordinate information in a symmetry-respecting manner. However, each encoder produces two outputs: updated scalar node features $\mathbf{h}$ and updated 3D coordinates $\mathbf{x}$. Only the scalar node features are retained as input to PRIME; the coordinate outputs are discarded. In EGNN, node features are updated via:
\begin{align*}
\mathbf{m}_{ij} &= \phi_e(\mathbf{h}_i, \mathbf{h}_j, \|\mathbf{x}_i - \mathbf{x}_j\|^2, a_{ij}), \\
\mathbf{m}_i &= \sum_{j \in \mathcal{N}(i)} \mathbf{m}_{ij}, \\
\mathbf{h}_i^{\text{out}} &= \phi_h(\mathbf{h}_i, \mathbf{m}_i),
\end{align*}
where $\|\mathbf{x}_i - \mathbf{x}_j\|^2$ is a rotation-invariant scalar squared distance and $\mathcal{N}(i)$ denotes the set of neighbors of node $i$. Since node features are updated solely through scalar quantities, $\mathbf{h}^{\text{out}}$ is invariant to rigid body transformations. The equivariant encoders are therefore used as invariant feature extractors: their internal equivariant message passing enables the computation of geometrically richer invariant scalar embeddings than standard invariant GNNs operating on scalar distances alone, while the retained outputs remain SE(3)-invariant.

\paragraph{Coordinate-Free Hierarchical GNN.} The hierarchical message passing framework operates entirely on scalar node feature vectors without accessing or updating 3D coordinates at any stage. All operations, including graph convolution, gated fusion, bottom-up aggregation via $(\Pi^{(\ell)})^\top$, and top-down refinement via $\Pi^{(\ell)}$, are coordinate-free linear transformations applied to scalar feature vectors. The hierarchical GNN is therefore SE(3)-invariant by construction.

\paragraph{End-to-End Invariance.} Since all input node features are SE(3)-invariant and the hierarchical GNN operates without 3D coordinates, the full PRIME framework achieves SE(3) invariance end-to-end. We note that the term ``Equivariant'' in PRIME refers to the use of equivariant encoders at the two finest-grained levels to extract geometrically expressive invariant features, rather than implying end-to-end equivariance of the full framework. SE(3) invariance is a desirable property for downstream protein function prediction, as protein function is independent of the arbitrary orientation of the protein in space.

\section{Dataset Details}
\label{app:datasets}

All datasets are sourced directly from ProteinWorkshop \cite{jamasb2024evaluating} using their preprocessed structures and standard splits. Table \ref{tab:dataset_stats} summarizes the key statistics for each benchmark.

\begin{table*}[ht]
\centering
\caption{Summary of benchmark datasets used for evaluation.}
\label{tab:dataset_stats}
\resizebox{\linewidth}{!}{
\begin{tabular}{llllcccc}
\toprule
Task & Level & Task Type & \#Train & \#Val & \#Test & \#Classes & Metric \\
\midrule
Fold Classification    & Graph & Multiclass Classification       & 12.3K & 0.7K & 1.3K / 0.7K / 1.3K & 1,195 & Accuracy \\
Reaction Class         & Graph & Multiclass Classification      & 29.2K & 2.6K & 5.6K               & 384   & Accuracy \\
Gene Ontology (BP)     & Graph & Multilabel Classification       & 27.5K & 3.1K & 3.0K               & 1,943 & $F_{\max}$ \\
Gene Ontology (MF)     & Graph & Multilabel Classification       & 27.5K & 3.1K & 3.0K               & 489   & $F_{\max}$ \\
Gene Ontology (CC)     & Graph & Multilabel Classification        & 27.5K & 3.1K & 3.0K               & 320   & $F_{\max}$ \\
PPI Site Prediction    & Node  & Binary Classification          & 478K  & 53K  & 117K               & 2     & ROC-AUC \\
\bottomrule
\end{tabular}
}
\end{table*}

\paragraph{Fold Classification.}
A multiclass graph classification task in which each protein is assigned to one of 1{,}195 fold classes derived from SCOP 1.75 \cite{hou2018deepsf}. Protein fold defines the global three-dimensional structure of a protein and is closely related to its function and stability. Predicting fold class from structure, therefore, serves as a fundamental benchmark for evaluating a model’s ability to capture global structural organization. The dataset provides three test splits of increasing difficulty (Family, Superfamily, and Fold) constructed by progressively reducing sequence similarity between training and test proteins, enabling assessment of structural generalization beyond sequence-level homology.

\paragraph{Reaction Class Prediction.}
A multiclass graph classification task where each protein is assigned to one of 384 Enzyme Commission (EC) reaction classes, defined using all four levels of the EC hierarchy \cite{hermosilla2021intrinsicextrinsic}. The dataset consists of experimentally resolved protein structures from the PDB and is split according to 50\% sequence identity, ensuring limited homology between training and test sets. Because enzyme function is governed by the geometry and physicochemical properties of catalytic sites, reaction class prediction evaluates a model’s ability to capture fine-grained functional information from protein structure.

\paragraph{Gene Ontology Prediction.}
A multilabel graph classification task in which each protein structure is annotated with Gene Ontology (GO) terms across three ontologies: Biological Process (BP), Molecular Function (MF), and Cellular Component (CC) \cite{gligorijevic2021structure}. The dataset is curated from experimentally determined PDB structures, with dataset splits based on a 30\% sequence similarity threshold. This task provides a comprehensive evaluation of protein function prediction, encompassing molecular activities, cellular localization, and high-level biological processes at varying levels of granularity.

\paragraph{PPI Site Prediction.}
A binary node-level classification task where individual residues are labeled according to whether they participate in a protein–protein interaction interface \cite{gainza2020deciphering}. The dataset is curated from experimental PDB structures, with interface residues defined based on inter-atomic distance criteria and standard train/validation/test splits preserved. In contrast to graph-level tasks, PPI site prediction requires accurate local structural representations at the residue level, directly assessing a model’s capacity to learn informative node embeddings for interaction-specific features.

\section{Additional Results}
\label{app:additional_results}

\subsection{Cross-Attention Readout on Reaction Class Prediction}

\begin{figure*}[ht]
    \centering
    \includegraphics[width=\linewidth]{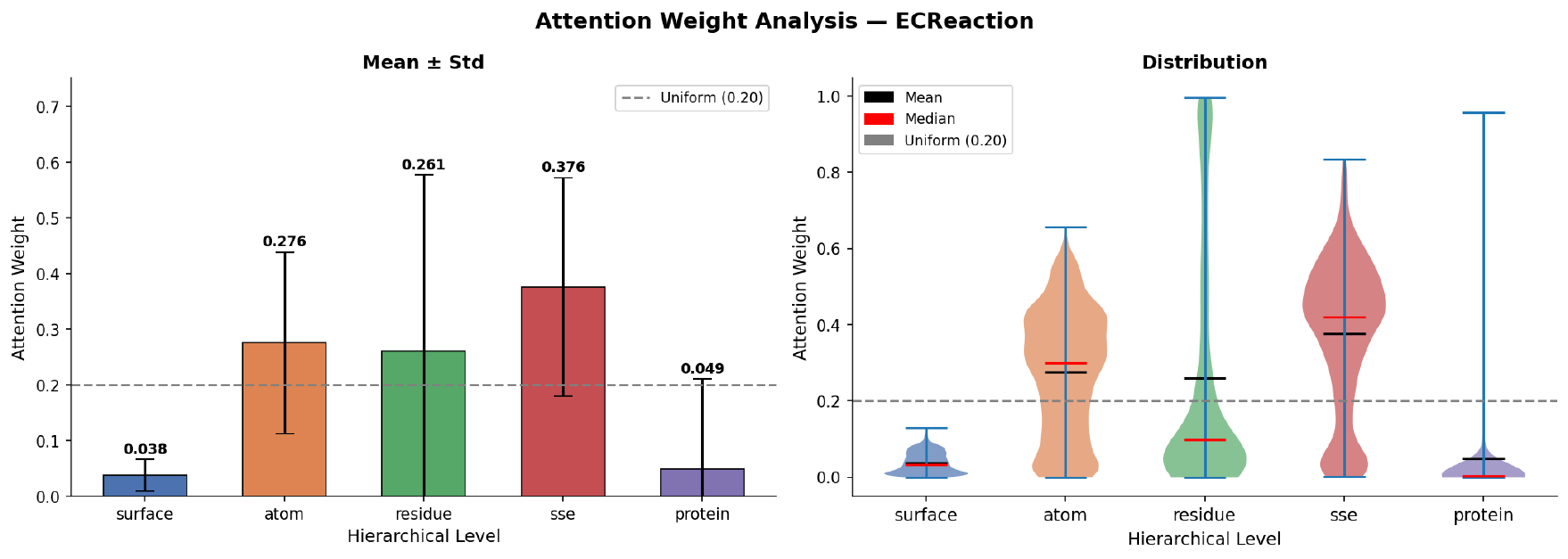}
    \caption{Learned cross-attention weights over structural levels for the adaptive cross-attention readout on Reaction Class prediction.}
    \label{fig:attn_reaction}
\end{figure*}

Figure \ref{fig:attn_reaction} presents the learned attention weights from the adaptive cross-attention readout applied to the Reaction Class prediction benchmark. Comparing these results with the Fold Classification analysis in the main text reveals meaningful differences in the model's structural preferences across tasks, providing empirical support for the hypothesis that level importance is task-dependent.

On the Reaction Class benchmark, the adaptive cross-attention readout achieves an accuracy of \textbf{85.02}, and the SSE level again receives the highest attention weight, while the atomic and residue levels receive substantially higher weights than in Fold Classification, both above the uniform baseline. The protein level, by contrast, receives lower attention than in Fold Classification. This may reflect the fact that global sequence context is less discriminative for reaction class, as proteins with very different sequences can catalyze the same reaction if they share similar active site geometry, making sequence-level information potentially less informative than structural detail at finer resolutions. The continued dominance of SSE is consistent with the known correlation between enzyme reaction class and protein fold topology; many EC classes are associated with characteristic secondary-structure arrangements that define the shape and accessibility of the catalytic pocket. The relatively higher attention on atomic and residue levels compared to Fold Classification may further reflect the importance of local geometric and chemical detail for distinguishing enzyme function, as catalytic activity depends on the precise spatial arrangement of residues and atoms within the active site. Surprisingly, the surface level receives the lowest attention weight at the readout stage, contrasting with its relatively higher weight in Fold Classification. We note, however, that this does not necessarily imply that surface information is uninformative for reaction class prediction; surface-level geometric features may have already been propagated upward to the atomic and residue levels through bottom-up message passing, with the model integrating surface information indirectly through finer-level representations rather than attending to it directly at prediction time.

These results provide direct empirical evidence that PRIME autonomously adjusts its structural preferences depending on the task. The contrasting attention patterns across Fold Classification and Reaction Class prediction validate the design choice of maintaining all five structural levels jointly within a unified hierarchy, as no single fixed resolution would be optimal across both tasks.

\end{document}